# Accountability of AI Under the Law: The Role of Explanation

*Finale Doshi-Velez,* Mason Kortz,* Ryan Budish, Chris Bavitz, Sam Gershman, David O'Brien, Kate Scott, Stuart Shieber, James Waldo, David Weinberger, Adrian Weller, Alexandra Wood*[1]

## I. INTRODUCTION

The use of applied machine learning for automated decision-making and decision assistance—popularly referred to as artificial intelligence or "AI"—has infused nearly every aspect of modern life to some degree. AI systems are currently used in applications ranging from automatic face-focus on cameras[2] and predictive policing[3] to segmenting MRI scans[4] and language translation.[5] They are being tested for safety-critical purposes such as clinical decision support[6] and autonomous driving.[7] However, AI systems continue to be poor at common sense reasoning and often fail at some tasks that are trivial for most people.[8] Moreover, decisions about how to define objective functions and what training data to use can introduce human error into AI decision making.[9] Thus, there exist legitimate concerns about the intentional and unintentional negative consequences of using AI systems.[10]

---

[1] This article is a product of over a dozen meetings of the Berkman Klein Center Working Group on AI Interpretability, a collaborative effort between legal scholars, computer scientists, and cognitive scientists. Authors would like to thank Elena Goldstein, Jeffrey Fossett, and Sam Daitzman for helping organize the meetings of the Working Group.

[2] Face detecting camera and method, U.S. Patent No. 6,940,545 (issued Sept. 6, 2005).

[3] Tong Wang, Cynthia Rudin, Daniel Wagner, and Rich Sevieri, *Learning to Detect Patterns of Ccrime*, *in* JOINT EUROPEAN CONFERENCE ON MACHINE LEARNING AND KNOWLEDGE DISCOVERY IN DATABASES 515 (2013).

[4] Abiodun M Aibinu, Momoh JE Salami, Amir A Shafie, and Athaur Rahman Najeeb, *MRI Reconstruction Using Discrete Fourier Transform: A Tutorial*, 42 WORLD ACADEMY OF SCIENCE, ENGINEERING AND TECHNOLOGY 179 (2008).

[5] Sunita Chand, *Empirical Survey of Machine Translation Tools*, 2 INTERNATIONAL CONFERENCE ON RESEARCH IN COMPUTATIONAL INTELLIGENCE AND COMMUNICATION NETWORKS 181 (2016).

[6] Amit X Garg, Neill KJ Adhikari, Heather McDonald, M Patricia Rosas-Arellano, PJ Devereaux, Joseph Beyene, Justina Sam, and R Brian Haynes, *Effects of Computerized Clinical Decision Support Systems on Practitioner Performance and Patient Outcomes: A Systematic Review*, 293 JOURNAL OF THE AMERICAN MEDICAL ASSOCIATION 1223 (2005).

[7] Markus Maurer, J Christian Gerdes, Barbara Lenz, and Hermann Winner, AUTONOMOUS DRIVING: TECHNICAL, LEGAL AND SOCIAL ASPECTS (2016).

[8] John McCarthy, PROGRAMS WITH COMMON SENSE (1960).

[9] Dario Amodei, Chris Olah, Jacob Steinhardt, Paul Christiano, John Schulman, and Dan Mané, *Concrete Problems in AI Safety* (arXiv preprint, arXiv:1606.06565 2016), *available at* https://arxiv.org/pdf/1606.06565.pdf.

[10] Nick Bostrom. *Ethical issues in Advanced Artificial Intelligence*, *in* SCIENCE FICTION AND PHILOSOPHY: FROM TIME TRAVEL TO SUPERINTELLIGENCE 277 (2003); D. Sculley, Gary Holt, Daniel Golovin, Eugene Davydov, Todd Phillips, Dietmar Ebner, Vinay Chaudhary, and Michael

To date, AI systems are only lightly regulated: it is assumed that AIs will be used for decision assistance and that the human user will use their common sense to make the final decision. However, even today we see many situations in which humans place too much trust in automated decision-making systems—consider the number of car accidents due to incorrect GPS directions,[11] or, at a larger scale, how incorrect modeling assumptions were at least partially responsible for the recent mortgage crisis.[12] As AI systems are used in more common and consequential contexts, there is increasing attention on whether and how they should be regulated.

One of the most commonly asked questions regarding AI regulation is "How can we take advantage of what AI systems have to offer while also holding AI developers and users accountable?" Accountability, in this context, means the ability to determine whether a decision was made in accordance with procedural and substantive standards and to hold someone responsible if those standards are not met.[13] The question of how to create accountable AI systems is important; accountability is an important element of good public and private governance.[14] It also a question that must be answered with some subtlety: poor choices may result in regulation that not only fails to truly improve accountability but also stifles the many beneficial applications of AI systems.[15]

While there are many tools for increasing accountability in AI systems, we focus on one in this report: explanation (we briefly discuss alternatives in Section 7). Explanations expose information about specific individual decisions without necessarily exposing the precise mechanics of the decision-making process. Explanations can be used to prevent or rectify errors and increase trust. Explanations can also be used to ascertain whether certain criteria were used appropriately or inappropriately in case of a dispute.

The question of when and what kind of explanation might be required of AI systems is urgent: details about a potential "right to explanation" were debated in the most recent revision of the European Union's General Data Protection Regulation (GDPR).[16] The resulting version of the GDPR, effective May 25, 2018, provides a right to information about the existence, logic, and envisaged consequences of automated decision-making systems in Articles 13 through 15, as well as a right not to be subject to

---

Young, *Machine Learning: The High-Interest Credit Card of Technical Debt*, SE4ML: SOFTWARE ENGINEERING 4 MACHINE LEARNING (NIPS 2014 WORKSHOP) (2014), *available at* https://ai.google/research/pubs/pub43146.

[11] Sarah Wolfe, *Driving into the Ocean and 8 Other Spectacular Fails as GPS Turns 25*, PUBLIC RADIO INTERNATIONAL (Feb. 17, 2014), https://www.pri.org/stories/2014-02-17/driving-ocean-and-8-other-spectacular-fails-gps-turns-25.

[12] Catherine Donnelly and Paul Embrechts, *The Devil is in the Tails: Actuarial Mathematics and the Subprime Mortgage Crisis*, 40 ASTIN BULLETIN: THE JOURNAL OF THE IAA 1 (2010).

[13] Joshua A Kroll, Solon Barocas, Edward W Felten, Joel R Reidenberg, David G Robinson, and Harlan Yu, *Accountable Algorithms*, 165 U. PA. L. REV. 633, 656 (2016).

[14] Jonathan Fox, *The Uncertain Relationship Between Transparency and Accountability*, 17 DEVELOPMENT IN PRACTICE 663, 663-65 (2007).

[15] Sandra Wachter, Brent Mittelstadt, and Luciano Floridi, *Transparent, Explainable, and Accountable AI for Robotics*, 2 SCIENCE ROBOTICS eaan6080 (2017), *available at* http://robotics.sciencemag.org/content/2/6/eaan6080/.

[16] Bryce Goodman and Seth Flaxman, *EU Regulations on Algorithmic Decision-Making and a 'Right to Explanation'*, AI MAGAZINE, Oct. 2017, at 50.

automated decision-making processes in Article 22.[17] The degree to which this constitutes a "right to explanation" is the subject of significant debate, but the fact that a major piece of legislation even arguably contains such a right prompts the question of whether explanations are necessary or sufficient to convey meaningful information about the operation of AI and other automated decision-making systems.

While the right to explanation in the ultimate version of the GDPR is ambiguous,[18] the issue of explainable AI has been noted by a number of public bodies in the United States and abroad. For example, in 2016, the U.S. Defense Advanced Research Projects Agency launched the Explainable AI program to fund research into "(1) how to produce more explainable models; (2) how to design the explanation interface; and (3) how to understand the psychological requirements for effective explanations."[19] In 2018, reports from the governments of the United Kingdom[20] and France[21] touched on the question of AI explainability. While there is significant support for explanations as a tool for holding AIs accountable, there are also concerns about the costs of generating explanations. In particular, there exist concerns that the engineering challenges surrounding explanation from AI systems would stifle innovation; that explanations might force trade secrets to be revealed; and that explanation would come at the price of system accuracy or other performance objectives.

This paper is a response to the recent debate over the role of explanations in improving the accountability of AI systems. In Section 2 of this document, we define what an explanation is and examine what kinds questions an explanation should answer. In Sections 3 and 4, we look at how explanations are used by society, specifically in U.S. and European legal and regulatory systems. We find that there is significant variation in how and when explanations are used, driven by factors such as the potential for harm, the possibility of correction or compensation, and the degree of suspicion that an error has or will be made. In Section 5, we describe technical considerations for designing AI systems to provide explanation while mitigating concerns about sacrificing prediction performance and divulging trade secrets. Under legally operative notions of explanations, AI systems are not indecipherable black-boxes; we can, and sometimes should, demand explanation from them. In Sections 6 and 7 we discuss the potential costs of requiring explanation from AI systems, situations in which explanation may not be appropriate, and finally other ways of holding AI systems accountable. Section 8 concludes with recommendations for future research.

[17] Council Regulation 2016/679, arts. 13-15, 22, 2016 O.J. (L119) 1.

[18] *See* Sandra Wachter, Brent Mittelstadt, and Luciano Floridi. *Why a Right to Explanation of Automated Decision-Making Does Not Exist in the General Data Protection Regulation*, 7 INTERNATIONAL DATA PRIVACY LAW 76, 79-83 (2017).

[19] DARPA, Broad Agency Announcement, *Explainable Artificial Intelligence (XAI)*, DARPA-BAA-16-53, at 6 (August 10, 2016), *available at* https://www.darpa.mil/attachments/DARPA-BAA-16-53.pdf.

[20] *See* House of Lords, Select Committee on Artificial Intelligence, Report of Session 2017-19, *AI in the UK: Ready, Willing, and Able?* (April 16, 2018), *available at* https://publications.parliament.uk/pa/ld201719/ldselect/ldai/100/100.pdf; House of Commons, Science and Technology Committee, Fourth Report of Session 2017-19, *Algorithms in Decision-Making* (May 15, 2018), *available at* https://publications.parliament.uk/pa/cm201719/cmselect/cmsctech/351/351.pdf

[21] *See* France Intelligence Artificielle, *Rapport de Synthèse* (Jan. 2017), *available at* https://www.economie.gouv.fr/files/files/PDF/2017/Rapport_synthese_France_IA_.pdf (French only); Cédric Villani, *For a Meaningful Artificial Intelligence: Towards A French and European Strategy* (March 28, 2018), *available at* https://www.aiforhumanity.fr/pdfs/MissionVillani_Report_ENG-VF.pdf.

## II. WHAT IS AN EXPLANATION?

In the colloquial sense, any clarifying information can be an explanation. Thus, we can "explain" how an AI makes operates in the same sense that we can explain how gravity works or explain how to bake a cake: by laying out the rules the system follows without reference to any specific decision (or falling object, or cake). However, access to the rules of a system—often referred to as "transparency"—may not be sufficient or even desirable for understanding AIs and holding them accountable.[22] The United Kingdom House of Lords, in a recent report, described the difference between transparency and explanation as follows:

> One solution to the question of intelligibility is to try to increase the technical transparency of the system, so that experts can understand how an AI system has been put together. This might, for example, entail being able to access the source code of an AI system. However, this will not necessarily reveal why a particular system made a particular decision in a given situation . . . An alternative approach is explainability, whereby AI systems are developed in such a way that they can explain the information and logic used to arrive at their decisions.[23]

In this paper, when we talk about an explanation for a decision, we mean a set of abstracted reasons or justifications for a particular outcome, not a description of the decision-making process in general. More specifically, we define the term "explanation" to mean human-interpretable information about the logic by which a decision-maker took a particular set of inputs and reached a particular conclusion.[24]

Furthermore, an explanation must also provide the correct *type* of information in order for it to be useful. As a governing principle for the content an explanation should contain, we offer the following: an explanation should permit a human observer to determine the extent to which a particular input was determinative or influential on the output. Another way of formulating this principle is to say that an explanation should be able to provide at least one of the following:

**Human-interpretable information about the factors used in a decision and their relative weight.** This is likely the most common understanding of what constitutes an explanation for a decision. A list of the factors that went into a decision, ideally ordered by the significance to the output, can provide accountability by confirming that proper procedures were followed. In many cases, society has prescribed a list of factors that must or must not be taken into account in a particular decision. For example, we many want to confirm that a child's interests were taken into account in a custody determination, or that race was not taken into account in a criminal prosecution. Even where there is no predetermined list of required or prohibited factors, a human-interpretable report on the factors used and their relative significance provides an opportunity to review and potentially contest the reasonableness of a decision.

[22] Kroll, *supra* note 13.

[23] House of Lords, *AI in the UK*, *supra* note 20, ¶¶ 95-100.

[24] Wachter, *Right to Explanation*, *supra* note 18. For a discussion about legibility of algorithmic systems more broadly, see Gianclaudio Malgieri and Giovanni Comande, *Why a Right to Legibility of Automated Decision-Making Exists in the General Data Protection Regulation*, 7 INTERNATIONAL DATA PRIVACY LAW 243 (2017).

**An answer to a counterfactual question.** In other cases, what we want to know is not just whether a factor was taken into account, but whether it was determinative of a specific outcome. This information can be obtained by posing counterfactual questions to the decision-maker.[25] For example, we may want to know what effect a particular change to an input has on the output or, conversely, what change must be made to the input to change the output in a particular way. Counterfactuals can also provide information about why two similar-looking sets of inputs resulted in different outputs, or vice versa. By isolating the determinative factors of a decision, which can then be contested, counterfactual explanations promote accountability. Counterfactuals also allow us to check for consistency between decisions, an important procedural element of accountability.

Having defined *what* constitutes an explanation, we next examine *when* explanations are desirable. In doing so, we lay the foundations for specific circumstances in which explanation are (or are not) currently required under the law (Section 4).

## III. SOCIETAL NORMS AROUND EXPLANATION

When it comes to human decision-makers, we often want an explanation when someone makes a decision we do not understand or believe to be suboptimal.[26] For example, was the conclusion accidental or intentional? Was it caused by incorrect information or faulty reasoning? The answers to these questions permit us to weigh our trust in the decision-maker and to assign blame in case of a dispute.

Society does not, however, demand an explanation for every suboptimal decision, for a number of reasons. First, explanations are not free. Generating them takes time and effort, thus reducing the time and effort available to spend on other, potentially more beneficial conduct. Therefore, the utility of explanations must be balanced against the cost of generating them. Another reason not to demand an explanation is the explanation might obscure more information than it reveals—humans are notoriously inaccurate when providing post-hoc rationales for decisions[27]— and even if an explanation is accurate, we cannot ensure that it will be used in a socially responsible way. Explanations can also change an individual's judgment: the need to explain a decision can have both positive and negative effects on the decision-maker's choices,[28] and access to an explanation might decrease observers' trust in some decisions.[29] Last but not least, social norms regarding individual autonomy weigh against demanding explanations for highly personal decisions.

[25] Sandra Wachter, Brent Mittelstadt, and Chris Russell, *Counterfactual Explanations Without Opening the Black Box: Automated Decisions and the GDPR* (arXiv preprint, arXiv:1711.00399, 2018), *available at* https://arxiv.org/ftp/arxiv/papers/1711/1711.00399.pdf.

[26] David Leake. *Evaluating Explanations: A Content Theory*. New York: Psychology Press, 1992.

[27] Richard E Nisbett and Timothy D Wilson, *Telling More Than We Can Know: Verbal Reports on Mental Processes*, 84 PSYCHOLOGICAL REVIEW 231 (1977); Chadd M Funk and Michael S Gazzaniga, *The Functional Brain Architecture of Human Morality*, 19 CURRENT OPINION IN NEUROBIOLOGY 678 (2009).

What, then, are the circumstances in which the benefits of an explanation outweigh the costs? We find that there are three factors that affect whether society considers a decision-maker to be morally, socially, or legally obligated to provide an explanation:

**The impact of the decision, especially the impact on persons other than the decision maker.** The more significant the effect of a decision is, the more likely society is to demand an explanation for it. However, even for important decisions, social norms generally will not compel an explanation for a decision that only affects the decision-maker, as doing so would unnecessarily infringe upon the decision-maker's independence. For example, if an individual invests their own funds and suffers losses, there is no basis to demand that the investor disclose their strategy. But if an investor makes a decision that loses a client's money, the client may well be entitled to an explanation.

**The possibility of contesting, correcting, or compensating for an error in the decision.** Society is more likely explanation where the explanation can be acted on in some way. This could mean overturning the decision, or assigning a blame and providing compensation for injuries caused by the decision. However, explanations can also be useful if they can positively change future decision-making or future behavior by the subject of the decision.[30] Conversely, if there is no recourse for the harm caused, then there is less justification for the cost of generating an explanation. For example, if a gambler wins a round of roulette, there is no reason to demand an explanation for the bet: there is no recourse for the casino and there is no benefit to knowing the gambler's strategy, as the situation is not repeatable.

**Reason to believe that an error has occurred (or will occur) in the decision-making process.** We are more likely to demand explanations when some element of the decision-making process—the inputs, the output, or the context of the process—conflicts with our expectation of how the decision will or should be made:

> *Unreliable or inadequate inputs.* In some cases, belief that an error has occurred arises from our knowledge of the decision-maker's inputs. An input might be suspect because we believe it is logically irrelevant. For example, if a surgeon refuses to perform an operation because of the phase of the moon, society might well deem that an unreasonable reason to delay an important surgery.[31] An input might also be forbidden. Social norms in the U.S. dictate that certain features, such as race, gender, and sexual identity or orientation, should not be taken into account deciding a person's access to employment, housing, and other social goods. If we know that a decision-maker has access to irrelevant or forbidden information—or a proxy for such information—it adds to our suspicion that the decision was improper. Similarly, there are certain features that we think *must* be taken into account for particular decision: if a person is denied a loan, but we know that the lender never checked the person's credit report, we might suspect that the decision was made on incomplete information and, therefore, erroneous.

---

[28] William F. Messier, Jr, William C. Quilliam, D. E. Hirst, and Don Craig, *The Effect of Accountability on Judgment: Development of Hypotheses for Auditing; Discussions; Reply* 11 AUDITING 123 (Supp. 1992).

[29] Jenny de Fine Licht, *Do We Really Want to Know? The Potentially Negative Effect of Transparency in Decision Making on Perceived Legitimacy,* 34 SCANDINAVIAN POLITICAL STUDIES 183 (2011).

[30] Wachter et al., *Counterfactual Explanations, supra* note 25.

*Inexplicable outcomes.* In other cases, belief that an error occurred comes from the output of the decision-making process, that is, the decision itself. If the same decision-maker renders different decisions for two apparently identical subjects, we might suspect that the decision was based on an unrelated feature, or even random. Likewise, if a decision-maker produces the same decision for two markedly different subjects, we might suspect that it failed to take into account a salient feature. Even a single output might defy our expectations to the degree that the most reasonable inference is that the decision-making process was flawed. If an autonomous vehicles suddenly veers off the road, despite there being no traffic or obstacles in sight, we could reasonably infer that an error occurred from that single observation.

*Interest in the integrity of the system.* Finally, we might demand an explanation for a decision even if the inputs and outputs appear proper because of the context in which the decision is made. This usually happens when a decision-maker is making highly consequential decisions and has the ability or incentive to do so in a way that is personally beneficial but socially harmful. For example, corporate directors may be tempted to make decisions that benefit themselves at the expense of their shareholders. Therefore, society may want corporate boards to explain their decisions, publicly and preemptively, even if the inputs and outputs of the decision appear proper.[32] Even when there is no reason to suspect that the decision-maker will act in a socially harmful way, explanations can increase trust in a system by providing proof that a decision was made according to a fair, robust, or accepted process.

We observe that the question of when it is reasonable to demand an explanation is more complex than identifying the presence or absence of these three factors. Each of these three factors may be present in varying degree, and no single factor is dispositive. When a decision has resulted in a serious and plainly redressable injury, we might require less evidence of improper decision-making. Conversely, if there is a strong reason to suspect that a decision was improper, we might demand an explanation for even a relatively minor harm. Moreover, even where these three factors are absent, a decision-maker may want to voluntarily offer an explanation as a means of increasing trust in the decision-making process.

## IV. EXPLANATIONS IN THE LAW

[31] Jean-Luc Margot, *No Evidence of Purported Lunar Effect on Hospital Admission Rates or Birth Rates*, 64 NURSING RESEARCH 168 (2015).

[32] Klaus J. Hopt, *Comparative Corporate Governance: The State of the Art and International Regulation*, 59 AM. J. COMP. L. 1, 6-16 (2011).

In the prior sections, we discussed in general terms the circumstances in which explanations are desirable. In this section, we turn to concrete examples in which explanations are not just desirable, but legally operative. While the substance of the law is subject to both interpretation and debate, it is still better defined than the moral, ethical, or social norms that govern when we *should* explain our actions. By focusing on the legal system, we narrow in on the most pressing circumstances—those where we *must* explain our actions.

We consider the role of explanations in the law from two perspectives and four countries: the United States, the United Kingdom, France, and Germany. The first perspective is of the legal system as a decision-making body; the second perspective is from the perspective of the legal system as enforcing accountability for decision-makers. While the following survey is far from comprehensive, it provides an overview of the importance of explanations under the law. Across both perspectives and countries, we find variations regarding the role of the explanation, who is obligated to provide it, and what type or amount of evidence is needed to trigger that obligation.

## A. Decision-Making in the Law

Legal adjudications are themselves a form of decision-making. Because of the significant potential impact of many legal decisions, there is a strong interest in holding legal decision-makers accountable. The role of explanation in providing such accountability varies based on the nature of both the decision and the decision-maker. To highlight this point, we consider the role of explanation in holding accountable two types of legal decision-makers: judges and juries.

**Explanation serves an important tool for accountability from judges.** In general, it is believed that judicial explanations help to guide and improve future decision-making, especially when the explanation is being generated by a higher court.[33] Judges are therefore required to generate explanations where the stakes of the case are high enough and there is a possibility of redress in the form of appellate review, even if there is no evidence that the judge has erred in the particular decision. Failure to give an adequate explanation of the reasons for a judicial decision can result in that decision being invalidated by a higher court.

This trend was consistent across all the jurisdictions we surveyed. In the United Kingdom, it is a common law principle that a judgment must be reasoned, meaning that it "explains to the parties and to any wider readership why the judge has reached the decision he has made."[34] If a judgment is not sufficiently explained, it can be vacated by a higher court. In France and Germany, the civil code expressly provides that all judgments must be reasoned.[35] Again, failure to provide an explanation may result in the judgment being vacated.

[33] *See* Frederick Schauer, *Giving Reasons*, 47 STAN. L. REV. 633, 641 (1995).

[34] *See, e.g.*, *Weymont v Place* [2015] EWCA Civ 289, [5].

However, each jurisdiction also admits to distinct variations and exceptions. One factor in whether or not an explanation is required is the impact of the decision. For example, under U.S. law, a judge ruling on an objection to testimony can do so with little or no explanation; the decision is highly discretionary.[36] Similarly, in the French legal system certain minor orders, such as an order to make a payment under an injunction, do not need to be reasoned.[37] Explanation might also be waived when a decision is highly personal and a public record of the reasons for the decision might constitute an unwarranted invasion of privacy. For example, French law also waives the reasoning requirement for some highly personal decisions, such as divorce and adoptions.[38] There are also situations where neither of these rationales clearly applies, yet unexplained decisions are nevertheless the norm, such as the United States Supreme Court's tradition of not giving reasons for denials of writs of certiorari.[39]

However, there are also instances where one might expect an explanation to be required where it is in fact not. For example, in the U.S. legal system, a judge handing down a criminal sentence—viewed as one of the most important decisions a court can make—traditionally did not have to give a reason for the sentence.[40] The practice of reviewing sentences for adequate explanation is a relatively recent invention.[41] Similarly, until recently, judges acting under the French Criminal Code did not need to provide reasoning for punishments. Only in 2018 did the French Supreme Court rule that a lack of explanation regarding a punishment violated the French Constitution.[42]

**Juries are often only required to offer limited explanations, if any.** In the United States and United Kingdom, juries are rarely required to explain their decisions at all. In fact, rules of procedure provide that juries generally cannot be interrogated as to the reasons for their decisions.[43] In France, jurors are required to answer a set of yes-or-no questions posed by the court, but are not required to explain how they reached these conclusions.[44] This is true despite the fact that a jury's deliberations may have an enormous impact on the parties to a case. One justification given for not demanding explanations from juries is that public accountability could bias jurors in favor of making popular but legally incorrect decisions.[45] Another is that opening jury decisions to challenges would weaken public confidence in the outcomes of trials and bog down the legal system with interminable retrials.[46] Explanations are not, therefore, widely used to hold juries accountable.

---

[35] *See* CODE DE PROCÉDURE CIVILE [C.P.C.] [CIVIL PROCEDURE CODE] art. 455 (Fr.); CODE DE PROCÉDURE PÉNALE [C. PÉN.] [CRIMINAL PROCEDURE CODE] art. 485 (Fr.); ZIVILPROZESSORDNUNG [ZPO] [CODE OF CIVIL PROCEDURE], § 313, para. 1, *translation at* https://www.gesetze-im-internet.de/englisch_zpo/englisch_zpo.html (Ger.); STRAFPROZESSORDNUNG [STPO] [CODE OF CRIMINAL PROCEDURE], § 34, *translation at* https://www.gesetze-im-internet.de/englisch_stpo/englisch_stpo.html (Ger.).

[36] Schauer, *supra* note 33, at 637.

[37] Cour de cassation [Cass.] [Supreme Court for Judicial Matters] 2e civ., May 16, 1990, Bull. civ. II, No. 103 (Fr.).

[38] CODE DE PROCÉDURE CIVILE [C.P.C.] [CIVIL PROCEDURE CODE] arts. 245-1, 353.

[39] Schauer, *supra* note 33, at 637.

[40] Michael M. O'Hear, *Appellate Review of Sentence Explanations: Learning from the Wisconsin and Federal Experiences*, 93 MARQ. L. REV. 751, 751-52 (2009).

[41] *See id.* at 753.

[42] Conseil constitutionnel [CC] [Constitutional Court], decision No. 2017-694 QPC, Mar. 2, 2018.

[43] Fed. R. Evid. 606(b); Contempt of Court Act 1981, c. 49 § 8 (UK).

[44] CODE DE PROCÉDURE PÉNALE [C. PÉN.] [CRIMINAL PROCEDURE CODE] arts. 3550-361-1.

That is not to say that juries are entirely black boxes. Juries in the United States or the United Kingdom can be required to return "special verdicts," meaning that instead of or in addition to finding for one party or the other, the jury must make findings on specific factual issues.[47] To the extent that special verdicts require the jury to confirm whether or not a specific factor was considered in reaching a conclusion, they constitute a form of explanation. Moreover, in specific circumstances, the rule that juries are not required to explain their decisions must give way to greater social concerns. For example, the United States Supreme Court recently held that if a juror makes a clear statement that they relied on race in reaching a decision, a court can consider that statement in deciding whether to grant a new trial.[48]

## B. Decision-Making Under the Law

In addition to rendering its own decisions, the legal system also passes judgment on the decisions of other parties, thereby providing legal accountability. Perhaps unsurprisingly, this sometimes requires the party being judged to generate an explanation for a decision. In the broadest sense, a party can be legally required to provide an explanation for a decision when the opposing party has provided some degree of proof that the decision caused a legal-cognizable and redressable injury. Beyond this general rules, particular laws and legal doctrines require parties to provide explanations in specific circumstances. We examine some of these circumstances here.

**Administrative agencies are legally required to explain their decisions as a matter of course.** In the United States, when an agency engages in rule-making, it must follow specific procedures that include generating a record of the reasons for the proposed or adopted rule. This record should include explanations as to how the agency resolved specific questions raised during the rulemaking process.[49] This record is explicitly linked to accountability: if an administrative rule is challenged in court, the reviewing judge will rely on the agency's record, not their own judgment, to determine whether the rule is proper. A rule that lacks an explanation will likely be struck down as arbitrary and capricious.[50]

Other jurisdictions we surveyed have similar rules. For example, in the United Kingdom, administrative acts can be challenged as irrational or procedurally unfair, among other things. In either case, the administrative body must provide an explanation for the challenged decision. Under the German Administrative Procedure Act, an agency that enacts a decision must provide an explanation that include the "chief material and legal grounds" for the decision.[51] However, this requirement is waived in specific circumstances, including when the act has a limited impact on the rights of individuals.[52] France has recently amended its administrative code with the Digital Republic Act, which creates a right for subjects of algorithmic decision-making by public entities to receive an explanation of the parameters (and their weighting) used in the decision-making process.[53]

[45] *See Clark v. United States*, 289 U.S. 1, 13, 53 S. Ct. 465, 469, 77 L. Ed. 993 (1933); *United States v. Symington*, 195 F.3d 1080, 1086 (9th Cir. 1999).

[46] *United States v. Thomas*, 116 F.3d 606, 618 (2d Cir. 1997).

[47] Kate H. Nepveu, *Beyond "Guilty" or "Not Guilty": Giving Special Verdicts in Criminal Jury Trials*, 21 YALE L. & POL'Y REV. 263, 269-80 (2003).

[48] *Pena-Rodriguez v. Colorado*, 137 S. Ct. 855, 869, 197 L. Ed. 2d 107 (2017).

[49] Jonathan Weinberg, *The Right to Be Taken Seriously*, 67 U. MIAMI L. REV. 149, 156-57 (2012).

[50] *Id.*

**Private decision-makers in certain industries can also be compelled to provide explanations for their decisions.** For example, in the U.S., the Fair Credit Reporting Act requires consumer reporting agencies to provide, with every request for a credit score, a list of the key factors that negatively influenced the consumer's score.[54] This provisions permits consumers to contest their credit scores, thereby adding a layer of accountability to the system. France, Germany, and the United Kingdom all operate under a comply-or-explain model of corporate governance. Under this model, private corporations must adhere to a corporate governance code. If they depart from the code in any way, they must publicly explain their reasons for doing so.[55]

**Finally, the legal system can be used to demand explanation from individual litigants on a case by case basis.** Across all of the jurisdictions we surveyed, adjudication of individual claims often requires one or both parties to generate explanations. Explanations are most common when civil or criminal liability turns on a defendant's state of mind, or *mens rea*. For example, an explanation for a particular choice can provide evidence as to whether the defendant acted knowingly, recklessly, negligently, or innocently—all of which can bear legal significance. Ordinarily, a defendant will not be compelled to explain a decision until the plaintiff or prosecutor has affirmatively established some evidence of wrongdoing or error. The precise amount of evidence required to compel an explanation varies with the governing law.

For example, in the United States, in a discrimination lawsuit under the Fourteenth Amendment, a plaintiff must provide some evidence that a decision made by the defendant—for example, a decision not to extend a government benefit to the plaintiff—was intentionally biased before the defendant is required to explain the decision.[56] But in certain circumstances, such as criminal jury selection, employment, or access to housing, evidence that the decision-making has a disparate impact on particular group is enough to shift the burden of explanation on the decision-maker.[57] In these cases, the historical prevalence of certain types of discrimination provides the basis for requiring explanations with less evidence of specific wrongdoing. A similar regime applies in the United Kingdom, where the burden is on the potentially discriminating party to prove explain why a decision with a discriminatory effect is nevertheless proportional and legitimate.[58]

---

[51] Verwaltungsverfahrensgeset [VwVfG] [Administrative Procedure Act], Jan 23, 2003, BGBL I at 102, last amended July 18, 2017, *translation at* http://www.wipo.int/wipolex/en/text.jsp?file_id=462505.

[52] *Id.*

[53] Loi 2016-1321 du 7 octobre 2016 pour une République numérique [Law 2016-1321 of Oct. 7, 2016 for a Digital Republic], JOURNAL OFFICIEL DE LA REPUBLIC FRANCAISE [J.O] [OFFICIAL GAZETTE OF FRANCE], Oct. 8, 2016, p. 235.

[54] Michael F. McEneney, Karl F. Kaufmann, *Implementing the Fact Act: Self-Executing Provisions*, 60 BUS. LAW. 737, 744 (2005).

[55] Commission Recommendation of 9 April 2014 on the Quality of Corporate Governance Reporting ('Comply or Explain'), 2014 O.J. (L 109) 43.

[56] David A. Strauss, *Discriminatory Intent and the Taming of Brown*, 56 U. CHI. L. REV. 935 (1989).

As the foregoing examples show, even in the relatively systematic and codified realm of the law, there are numerous factors that affect whether human decision-makers will be required to explain their decisions. These factors include the nature of the decision, the susceptibility of the decision-maker to outside influence, moral and social norms, the perceived costs and benefits of an explanation, and a degree of historical accident. The varying emphasis on explanation also reflects the multiple roles legal systems play within their countries and around the world. Where the focus is on individual responsibility and restitution, such as with criminal law and personal injury liability, individual explanations are often central to legal outcomes. Where the law is more concerned with social welfare, there may be a greater emphasis on empirical evidence (as with certain anti-discrimination laws) or process guarantees (as with administrative rule-making).

## V. IMPLICATIONS FOR AI SYSTEMS

With our current legal contexts in mind, we now turn to technical considerations for extracting explanation from AI systems. Is it possible to create AI systems that provide the same kinds of explanation that are currently expected of humans, in the contexts that are currently expected of humans, under the law? An answer to the affirmative implies a simple way to handle, with minimal changes to the law, situations in which AIs make decisions currently performed by humans: we can ask of AIs the same that we ask of humans. (Of course, as AI technologies mature, the law can and should adapt to have different standards specific to AIs and standards specific to humans—see Sections 1 and Section 7.)

**Legally-Operative Explanations are Feasible.** Modern machine learning systems commonly have millions of parameters, suggesting that they are impossible to explain. However, there exists an important distinction between transparency—knowing exactly how a system behaves—and a legally-operative explanation—which must only assist in answering the kinds of questions in Section 2. Specifically, neither identifying important factors nor reasoning about their counterfactuals requires knowing the flow of bits through an AI system, no more than explanation from humans requires knowing the flow of signals through neurons (which would also be uninterpretable to a human!). In fact, quantifying how changes in inputs impact outputs is a well-studied problem in the statistical and AI literature, and the two ideas above, identifying the important factors and reasoning about their counterfactuals, can be mapped to two technical concepts in AI: *local explanation* and *local counterfactual faithfulness*.

[57] Joel H. Swift, The Unconventional Equal Protection Jurisprudence of Jury Selection, 16 N. ILL. U. L. REV. 295 (1996); Justin D. Cummins & Beth Belle Isle, Toward Systemic Equality: Reinvigorating A Progressive Application of the Disparate Impact Doctrine, 43 MITCHELL HAMLINE L. REV. 102 (2017).

[58] Equality Act 2010, c. 15, s. 19 (Gr. Brit.).

> *Answering Questions about Important Factors and their Weight: Local Explanation.* In the AI world, explanation is often formalized as a (perhaps simplified) rule that describes the decision-making process in terms that a human can understand (else it defeats the purpose). One common way to simplify an explanation is not to try to explain the *global* behavior of the system—that is, how it will make decisions in all circumstances—but only the *local* behavior relevant for a particular input.[59] Here, locality implies that the important factors may be different for different instances. For example, for one person, payment history may be the reason behind their loan denial, for another, insufficient income. This notion of locality directly maps to the notion of explaining a specific decision, which is the most common case of when explanation is required under the law.
>
> Local behaviors are much easier to characterize than global ones. AI systems are naturally designed to have specific inputs varied, differentiated, and passed through many other kinds of computations—all in a reproducible and robust manner. It is already the case that AI systems are trained to have relatively simple local decision boundaries to improve prediction accuracy, as we do not want tiny perturbations of the input changing the output in large and chaotic ways.[60] Thus, we can readily expect to answer the first question in Section 2—what were the important factors in a decision—by systematically probing the inputs to determine which have the greatest effect on the outcome.

---

[59] Marco Tulio Ribeiro, Sameer Singh, and Carlos Guestrin, *"Why Should I Trust You?": Explaining the Predictions of Any Classifier*, 22 ACM SIGKDD INTERNATIONAL CONFERENCE ON KNOWLEDGE DISCOVERY AND DATA MINING 1135 (2016); Tao Lei, Regina Barzilay, and Tommi Jaakkola, *Rationalizing Neural Predictions* (arXiv Preprint, arXiv:1606.04155) (2016); Philip Adler, Casey Falk, Sorelle A Friedler, Gabriel Rybeck, Carlos Scheidegger, Brandon Smith, and Suresh Venkatasubramanian, *Auditing Black-Box Models for Indirect Influence*, 16 INTERNATIONAL CONFERENCE ON DATA MINING 1 (2016); Ruth Fong and Andrea Vedaldi, *Interpretable Explanations of Black Boxes by Meaningful Perturbation* (arXiv preprint, arXiv:1704.03296) (2017); Ramprasaath R Selvaraju, Abhishek Das, Ramakrishna Vedantam, Michael Cogswell, Devi Parikh, and Dhruv Batra, *Grad-cam: Why Did You Say That? Visual Explanations from Deep Networks via Gradient-based Localization* (arXiv preprint arXiv:1610.02391) (2016); Daniel Smilkov, Nikhil Thorat, Been Kim, Fernanda Viégas, and Martin Wattenberg, *Smoothgrad: Removing Noise by Adding Noise,* (arXiv preprint, arXiv:1706.03825) (2017); Avanti Shrikumar, Peyton Greenside, Anna Shcherbina, and Anshul Kundaje, *Not Just a Black Box: Interpretable Deep Learning by Propagating Activation Differences* (arXiv preprint, arXiv:1704.02685) (2016); Pieter-Jan Kindermans, Kristof T Schütt, Maximilian Alber, Klaus-Robert Müller, and Sven Dähne, *Pattern-net and Ppatternlrp – Improving the Interpretability of Neural Networks* (arXiv preprint, arXiv:1705.05598) (2017); Andrew Ross, Michael C Hughes, and Finale Doshi-Velez, *Right for the Right Reasons: Training Differentiable Models by Constraining Their Explanation*, 26 INTERNATIONAL JOINT CONFERENCE ON ARTIFICIAL INTELLIGENCE 2662 (2017); Sameer Singh, Marco Tulio Ribeiro, and Carlos Guestrin, *Programs as Black-box Explanations* (arXiv preprint, arXiv:1611.07579) (2016).

[60] Harris Drucker and Yann Le Cun, *Improving Generalization Performance using Double Backpropagation*, 3 IEEE TRANSACTIONS ON NEURAL NETWORKS 991 (1992); Kevin P Murphy, MACHINE LEARNING: A PROBABILISTIC PERSPECTIVE (2012).

> *Answering Questions about Counterfactuals: Counterfactual Faithfulness.* Counterfactual faithfulness allows us to answer the remaining questions from Section 2: whether a certain factor determined the outcome, and related, what factor caused a difference in outcomes. For example, if a person was told that their income was the determining factor for their loan denial, and then their income increases, they might reasonably expect that the system would now deem them worthy of getting the loan. Importantly, however, we only expect that counterfactual faithfulness apply for related situations—we would not expect an explanation in a medical malpractice case regarding an elderly, frail patient to apply to a young oncology patient. However, we may expect it to still hold for a similar elderly, less frail patient. Recently, Wachter et al. also pointed out how counterfactuals can provide the cornerstone for explanations in many situations.[61]

Importantly, both of these properties above can be satisfied *without* knowing the details of how the system came to its decision. For example, suppose that the legal question is whether race played an inappropriate role in a loan decision. One might then probe the AI system with variations of the original inputs changing only the race. If the outcomes were different, then one might reasonably argue that race played a role in the decision. And if it turns out that race played an inappropriate role, that constitutes a legally sufficient explanation—no more information is needed under the law (although the company may internally decide to determine the next level of cause, e.g. bad training data vs. bad algorithm). This point is important because it mitigates concerns around trade secrets: explanation can be provided without revealing the full internal contents of the system.

**Mapping inputs and intermediate representations in AI systems to human-interpretable concepts will be challenging.** While the properties above allow us to map properties of legally-operative explanations to technical definitions, there remains a key technical challenge of *concept-mapping*. AIs do not see the world as humans do: they often take in large collections of variables—pixel values, hospital codes, purchasing histories—without any understanding of how these variables might related to human-interpretable terms such as race or gender. For example, self-driving cars may have multitudes of sensors, each with high-dimensional range and vision inputs; the human brain already converts its visual inputs into higher-level concepts such as trees or street signs. Clinical decision support systems may take in tens of thousands of variables about a patient's diagnoses, drugs, procedures, and concepts extracted from the clinical notes; the human doctor has terms like sepsis or hypertension to describe constellations of these variables. Thus, answering a question like "Did race play a determinative role in the decision?" may not be as simple as adjusting the "race" input.

While there do exist methods to map the high-dimensional inputs to an AI system to human-interpretable concepts, the process generally requires training the system with large amounts of data in which both the raw input and the associated concept are given. As such, explanations from AI systems will be most straight-forward if the relevant terms are known in advance. In this case, the AI system can be trained to map its inputs to the relevant terms. For example, in the medical sphere, there are a number of algorithms for determining whether a patient has diabetes from a multitude of inputs;[62] recent work has identified ways to weigh the importance of much more general terms.[63] These terms can then be used in constructing the explanation, rather than the raw inputs (we emphasize the more than those terms might be used in the decision, but we can only answer questions with respect to those terms). There will be some technical innovation required, but by and large we see relatively few difficulties for AI systems to provide the kinds of explanation that are currently required in the case where legislation or regulation makes it clear what terms may be asked for *ex ante*; there is also an established process for companies to adapt new standards as legislation and regulation change.

[61] Wachter, *Counterfactual Explanations*, *supra* note 25.

That said, even in the *ex ante* case, there are subtleties. While it is relatively straightforward to identify what inputs are correlated with certain terms, and verify whether predictions of terms are correlated with decisions, it will require some work to determine ways to test counterfactuals. For example, how can we show that a security system that uses images of a face as input does not discriminate against gender? One would need to consider an alternate face that was similar in every way except for gender. How would a car manufacturer know if its AI system 'saw' an ice patch in the road? One would need to consider an alternate set of inputs in which the ice patch was removed from all the system's sensors. Still, we believe that these are technical challenges that can be overcome.

Another subtlety is that, to create the required terms, the AI system will need access to potentially sensitive information. Currently, we often assume that if the human did not have access to a particular term, such as race, then it could not have been used in the decision. However, it is very easy for AI systems to reconstruct sensitive information from high-dimensional inputs. Data about shopping patterns can be used to identify a person's age, gender, and socio-economic status, as can data about healthcare utilization. Especially with AI systems, excluding a protected category does not mean that a proxy for that category is not being created. Thus, a corollary to the arguments above is that we must measure any terms that we wish to protect against, to be able to ensure that we are not generating proxies for them. Our legal system must allow them to be collected, and AI system designers should build ways to test whether systems are creating that term and using it inappropriately. Regulation must be put in place so that any protected terms collected by AI system designers are used only to ensure that the AI system is designed correctly, and not for other purposes within the organization. (It would be unfortunate, to say the least, if we can verify that an AI system is fair, only to find that a human decision-maker is accessing forbidden information and biasing the final decision.)

---

[62] Katherine M Newton, Peggy L Peissig, Abel Ngo Kho, Suzette J Bielinski, Richard L Berg, Vidhu Choudhary, Melissa Basford, Christopher G Chute, Iftikhar J Kullo, Rongling Li, et al., *Validation of Electronic Medical Record-based Phenotyping Algorithms: Results and Lessons Learned from the Emerge Network*, 20 JOURNAL OF THE AMERICAN MEDICAL INFORMATICS ASSOCIATION e147 (2013).

[63] Been Kim, Martin Wattenberg, Justin Gilmer, Carrie Cai, James Wexler, Fernanda Viegas, and Rory Sayres, *Interpretability Beyond Feature Attribution: Quantitative Testing with Concept Activation Vectors (TCAV)*, 2018 International Conference on Machine Learning 2673.

The challenges increase if the relevant terms are only determined *ex post* —such as in litigation scenarios. In such cases, AI systems may struggle; unlike humans, they cannot be asked to refine their explanations after the fact without additional training data. For example, we cannot identify what proxies there are for age in a data set if age itself has never been measured. For such situations, we first note that there is precedent for what to do in litigation scenarios when some information is not available, ranging from drawing inferences against the party that could have provided the information to imposing civil liability for unreasonable record-keeping practices.[64] Second, while not always possible, in many cases it may be possible to quickly train a proxy—especially if AI designers have designed the system to be updated—or have the parties mutually agree (perhaps via a third party) what are acceptable proxies. The parties may also agree to assessment via non-explanation-based tools.

**Explanation systems should be considered distinct from AI systems.** Finally, we provide guidance on how to think about the process by which an AI produces an explanation. We suggest that regulation around explanation from AI systems should consider the explanation system as *distinct* from the AI system. Figure 1 depicts a schematic framework for explainable AI systems. The AI system itself is a (possibly proprietary) black-box that takes in some inputs and produces some predictions. The designer of the AI system likely wishes the system's predictions ($\hat{y}$) to match the real world ($y$). The designer of the *explanation system* must output a *human-interpretable* rule $e_x()$ that takes in the same input $x$ and outputs a prediction $\tilde{y}$. To be locally faithful under counterfactual reasoning formally means that the predictions $\tilde{y}$ and $\hat{y}$ are the same under pre-defined perturbations of the input $x$. This description also emphasizes the difference between the rule from the explanation system—which is a simplified version of the AI, accurate only near an input of interest—and the full system, which may be far more complex.

*Figure 1: Diagram of a Framework for Explainable AI Systems*

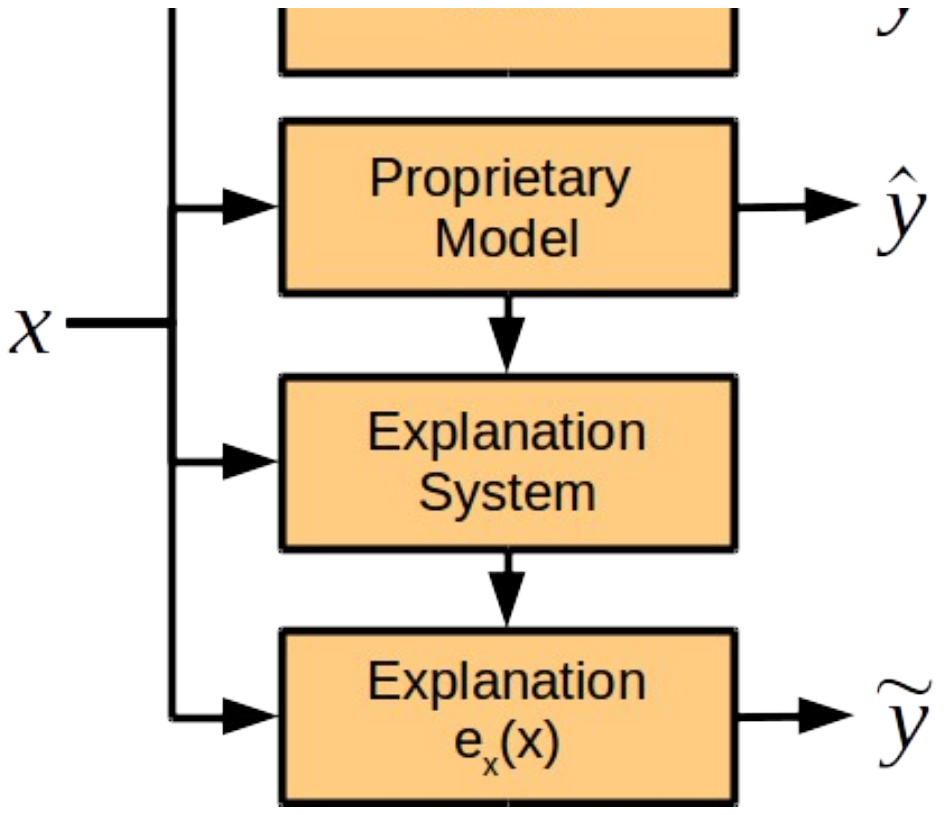

This framework renders concepts such as local explanation and local counterfactual faithfulness readily quantifiable. For any input $x$, we can check whether the prediction made by the local explanation ($\tilde{y}$) is the same as the prediction made by the AI system ($\hat{y}$). We can also check whether these predictions remain consistent over desired perturbations of $x$ (e.g. changing the race). Thus, not only can we measure what proportion of the time an explanation system is faithful, but we can also identify the specific instances in which it is not. From a regulatory perspective, this opens the door to regulation that requires that an AI system be explainable some proportion of the time or in certain kinds of contexts—rather than all the time. Loosening the explanation requirement in this way may allow for the AI system to use a much more complex logic for a few cases that really need it.

[64] Steffen Nolte, The Spoliation Tort: An Approach to Underlying Principles, 26 ST. MARY'S L.J. 351 (1995), Michael Cicero, Drug Testing of Federal Government Employees: Is Harm Resulting from Negligent Record Maintenance Actionable?, 1988 U. CHI. LEGAL F. 239 (1988).

More broadly, thinking of an explanation system as distinct from the original AI system also creates opportunities for industries that specialize in explanation systems. It also highlights where the true cost of demanding explanations lies: as a separate system, the explanation system will *not* affect the accuracy of the original predictor. However, depending on the application, *building* such a system may have nontrivial cost. Finally, there is a question of what it means for the explanation to be *human-interpretable*, especially when humans themselves are notorious for both providing manipulative explanations and misunderstanding explanation from fellow humans. Here, we emphasize that at least explanations from AIs are at least not creating new challenges; concerns about whether the explanation will be correctly utilized must be addressed for AI-generated explanations just as they must be addressed for human-generated explanations.

In summary, the machine learning community already has the technical frameworks in place to be able to provide legally-operative explanations. To build AI systems that can provide explanation in terms of human-interpretable terms, we must both list those terms and allow the AI system access to examples to learn them. System designers should design systems to learn these human-interpretable terms, and also store data from each decision so that is possible to reconstruct and probe a decision post-hoc if needed. Policy makers should develop guidelines to ensure that the explanation system is being faithful to the original AI.

## VI. COMPARISON OF HUMAN AND AI CAPABILITY FOR EXPLANATION

So far, we have argued that legally-operable explanation from AI is technically feasible in the situations the law requires explanation from humans. This approach would prevent otherwise legally accountable decision-makers from "hiding" behind AI systems, while not requiring the developers of AI systems to spend resources or limit system performance simply to be able to generate legally unnecessary explanations. While there are some technical challenges to be overcome, we believe it is possible to build explanation systems that do not impact the predictive performance of an AI, that simply explain the AI that has already been built.

That said, there are obviously salient differences between AI systems and humans. Should this affect the extent to which AI explanations should be the subject of regulation? There may be situations in which it is possible to demand more from humans, and other situations in which it might be possible to hold AI systems to a higher standard of explanation. More broadly, there may also be situations in which we currently achieve accountability via explanation, but could more efficiently achieve it via other means for AI systems. In this section, we describe the differences in the capability for humans and AIs to generate explanation, and we discuss broader alternatives in Section 7.

Given the myriad variations in when explanations are currently required (as described in Section 4), we cannot hope to analyze how humans and AIs might perform in each. At the most general level, though, we can categorize the factors that relate to the use of explanation as either extrinsic or intrinsic to the decision-maker. Extrinsic factors—the significance of the decision, the relevant social norms, the extent to which an explanation will inform future action—are likely to be the same whether the decision-maker is a human or an AI system. Extrinsic factors do not depend on human or AI capabilities. However, intrinsic factors—capabilities—vary significantly between humans and AIs (see Table 1), and will likely be key in eventually determining where demands for human and AI explanations under the law should overlap and where they should diverge.

One important intrinsic difference between AIs and humans is that AIs must prepare to provide explanation *ex ante*, while we generally assume that humans will, in the course of making a decision, store the information required to produce an explanation *ex post*. For example, a doctor who does not explain the reasons for a diagnosis at the time it is made can nevertheless provide those reasons if, after the fact, diagnosis is incorrect and they get sued. Sometimes, a decision-maker might be required to create a record to aid in the subsequent generation of an explanation—for example, many medical providers require doctors to annotate patient visits for this very reason, despite the fact that it takes extra time. However, requiring human decision-makers to document their decisions is the exception, not the norm. Therefore, the costs and benefits of generating a human explanation can be assessed at the time the explanation is requested. Moreover, a human decision-maker is able to draw on the totality of their experience to produce a relevant and helpful explanation tailored to a specific use case.

In contrast, AI systems do not automatically store information about their decisions. Often, this feature is considered an advantage: unlike human decision-makers, AI systems can delete information to optimize their data storage and protect privacy. However, an AI system designed this way would not be able to generate *ex post* explanations the way a human can. Instead, whether resources should to be allocated to explanation generation becomes a question of system design. This is analogous to the question of whether a human decision-maker should be required to keep a record. The difference is that with an AI system this design question must *always* be addressed explicitly. Given the constraints of current AI technology, even the most robust explanation system will lack the ability of a human to tailor its explanations to a given set of circumstances.

Another important intrinsic difference is that AIs have perfect memory and do not suffer from cognitive biases and social pressure. If an AI system is designed to store information about a decision (rather than delete it), then the inputs, all intermediate steps, and the final outputs can be stored exactly (although transparency may be required to verify this). Therefore, they do not suffer from the cognitive biases that can make human explanations unreliable; they cannot be tricked or pressured into providing an explanation that is more convenient but not accurate. Additionally, unlike humans, AI systems are not vulnerable to the social pressures that could alter their decision-making processes. Thus, there is no need to shield AI systems from generating explanations, for example, the way the law shields juries. However, there may be other valid reasons to limit generation of, or at least access to, explanations for AI decisions. For example, access to a sufficiently high number of explanations from a given system could allow the system to be reverse-engineered, exposing intellectual property, flaws in the system, or sensitive data.[65]

| | **Human** | **AI** |
|---|---|---|
| *Strengths* | Can provide explanation post-hoc, can adapt to specific circumstances | Reproducible, no cognitive biases, no social pressure |
| *Weaknesses* | May be inaccurate and unreliable, feel social pressure | Inflexible, requires up-front engineering, can be reverse engineered |

*Table 1: Comparison of Human and AI Capabilities for Explanation*

## VII. ALTERNATIVES TO EXPLANATION

As described in in Section 4, explanation is one of many tools used to assign rights and liabilities in the legal system. Similarly, explanation is but one way to hold AI systems accountable. As AI technologies mature, it may turn out that certain kinds of errors from AI decisions are best considered 'product defects' under the law, while others are best considered 'toxic leaks'—and only some are considered akin to the 'human decisions' that require explanation. In this section, we discuss the trade-offs associated with three core classes of accountability tools for AIs: explanation, empirical evidence, and theoretical guarantees.

*Explanation.* Explanations are our one of our key tools when determining fault for an individual event. While we argued in Section 5 that legally-operable definitions are technically feasible, there may still exist reasons that we prefer an alternative way to achieve accountability. From a technical perspective, an explanation system may struggle if a new factor is suddenly needed. Designing a system to also provide explanation is a non-trivial engineering task, and thus requiring explanation may create a financial burden that disadvantages smaller companies. In some cases, the need to provide a simple-enough explanation may impact the accuracy of a system, and one must decide if the additional errors made by the AI are outweighed by the errors caught by the explanation system. Similar situations may occur even if the AI is not designed to reject solutions that fall below a threshold of explicability; the human responsible for implementing the solution may discard a solution with a more complex explanation for a solution with a more appealing—or legally defensible—explanation. Given that one of the purported benefits of AI decision-making is the ability to identify patterns that humans cannot, this situation would be counterproductive. For some of these cases, we might, as a society, that certain individual outcomes require compensation without an assignment of blame or innocence—for example, a product defect.

*Empirical Evidence.* In many situations, we do not need to assign blame or innocence for an individual decision—a measure of a system's overall performance is sufficient to justify (or implicate) a decision-making system by demonstrating the value (or harm). For example, we might observe that an autonomous aircraft landing system has fewer safety incidents than human pilots, or that the use of a clinical diagnostic support tool reduces mortality. As recognized in many U.S. anti-discrimination statutes, questions of bias or discrimination are often best ascertained statistically: for example, a loan approval system might demonstrate its bias by approving more loans for men than women when other factors are controlled for. In fact, in some cases statistical evidence is the only kind of justification that is possible; certain types of subtle errors or discrimination may only show up in aggregate. Empirical evidence can also be used to determine whether errors even out over an individual. While empirical evidence is not unique to AI systems, AI systems, as digesters of big data used in highly reproducible ways, are particularly well-suited to provide empirical evidence.

[65] Smitha Milli, Ludwig Schmidt, Anca D Dragan, and Moritz Hardt, *Model Reconstruction From Model Explanations,* (arXiv preprint arXiv:1807.05185) (2018).

*Theoretical Guarantees.* In rare situations, we might be able to provide theoretical guarantees about an AI system. For example, we trust our encryption systems because they are backed by proofs; neither explanation nor evidence are required. Similarly, if there are certain agreed-upon schemes for voting and vote counting, then it may be possible to design a system that provably follows those processes. Likewise, a lottery is shown to be fair because it abides by some process, even though there is no possibility of fully explaining the generation of the pseudo-random numbers involved. This is like the principle of administrative law that once a process has been established as adequate, any output of that process is presumed to be substantively correct. With sufficiently transparent AIs, we can do more than presume that the established process was followed; in some circumstances, we can prove it. Theoretical guarantees are a form of perfect accountability that only AI systems can provide (as long as they are properly implemented); however, these guarantees require very cleanly specified contexts that often do not hold in real-world settings.

We emphasize that the trade-offs associated with all of these methods will shift as technologies change. For example, access to greater computational resources may reduce the computational burden associated with explanation, but enable even more features to be used, increasing the challenges associated with accurate summarization. New modes of sensing might allow us to better measure safety or bias, allowing us to rely more on empirical evidence, but they might also result in companies deciding to tackle even more ambitious, hard-to-formalize problems for which explanation might be the only available tool. We summarize considerations for choosing an accountability tool for AI systems in Table 2.

| **Approach** | **Well-Suited Contexts** | **Poorly-Suited Contexts** |
|---|---|---|
| *Theoretical Guarantees* | Situations in which both the problem and the solution can be fully formalized (gold standard, for such cases) | Any situation that cannot be sufficiently formalized (most cases) |
| *Statistical evidence* | Problems in which outcomes can be completely formalized, and we take a strict liability view; problems where we can wait to see some negative outcomes happen so as to measure them | Situations where the objective cannot be fully formalized in advance |
| *Explanation* | Problems that are incompletely specified, where the objectives are not clear and inputs might be erroneous | Situations in which other forms of accountability are possible |

*Table 2: Considerations for Approaches for Holding AIs Accountable*

## VIII. RECOMMENDATIONS

In the sections above, we have discussed the circumstances in which humans are required to provide explanation under the law, as well as what those explanations are expected to contain. We have also argued that it should be technically feasible to create AI systems that provide the level of explanation that is currently required of humans. The question, of course, is whether we *should*. The fact of the matter is that AI systems are increasing in capability at an astounding rate, with optimization methods of black-box predictors that far exceed human capabilities. Making such quickly-evolving systems be able to provide explanation, while feasible, adds an additional amount of engineering effort that might disadvantage less-resourced companies because of the additional personnel hours and computational resources required; these barriers may in turn result in companies employing suboptimal but easily-explained models.

Thus, just as with requirements around human explanation, we will need to think about why and when explanations are useful enough to outweigh the cost. Requiring every AI system to explain every decision could result in less efficient systems, forced design choices, and a bias towards explainable but suboptimal outcomes. For example, the overhead of forcing a toaster to explain why it thinks the bread is ready might prevent a company from implementing a smart toasting feature—either due to the engineering challenges or concerns about legal ramifications. On the other hand, we may be willing to accept the monetary cost of an explainable but slightly less accurate loan approval system for the societal benefit of being able to verify that it is nondiscriminatory. As discussed in Section 3, there are societal norms around when we need explanation, and these norms should be applied to AI systems as well.

For now, we posit that demanding explanations from AI systems is reasonable, and that we should start by asking of our AI systems what we ask of humans. Doing so avoids AI systems from getting a “free pass” to avoid the kinds of scrutiny that may come to humans, and also avoids asking so much of AI systems that it would hamper innovation and progress. Even this modest step will have its challenges, and as they are resolved, we will gain a better sense of whether and where demands for explanation should be different between AI systems and humans. As we have little data to determine the actual costs of requiring AI systems to generate explanations, the role of explanation in ensuring accountability must also be re-evaluated from time to time, to adapt with the ever-changing technology landscape.